\documentclass[a4paper]{article}
\usepackage[german,english]{babel}
\usepackage[T1]{fontenc}
\usepackage{ae,aecompl}
\usepackage{amsfonts}
\usepackage{amsmath}
\usepackage{amsthm,amssymb}
\usepackage{float}
\usepackage{times}
\usepackage{algorithm}
\usepackage{algpseudocode}
\usepackage{graphicx}
\usepackage{epstopdf}
\usepackage{enumitem}
\usepackage{array}

\makeatletter
\renewcommand{\ALG@beginalgorithmic}{\small}
\makeatother

\setlist[itemize]{noitemsep}

\newfloat{algorithm}{thp}{lop}
\floatname{algorithm}{Algorithm}

\RequirePackage{color}
\RequirePackage{soul}
\setstcolor{blue}

\usepackage{graphicx}
\usepackage{stfloats}
\DeclareGraphicsExtensions{.pdf,.ai,.eps,.ps,.jpg}
\graphicspath{{./figures/}} 

\newcommand{\bsfigure}[4][tb]{%
  \begin{figure}[#1]
    \centering
    \includegraphics[scale=#2]{#3}
    \vglue 0ex plus 0.5ex minus 3.5ex
    {\caption{\small #4.}\label{#3}}%
  \end{figure}}

\title{A Novel Anomaly Detection Algorithm for Hybrid Production Systems based on Deep-Learning and Timed Automata}

\usepackage{booktabs}
\usepackage{xspace}

\newcommand{\nat}{\ensuremath{\mathbf{N} }}

\newcommand{\real}{\ensuremath{\mathbf{R} }}

\newcommand{\dad}{\textit{DAD:DeepAnomalyDetection}}

\newtheorem{definition}{Definition}

\title{A Novel Anomaly Detection Algorithm for Hybrid Production Systems based on Deep Learning and Timed Automata}

\author%
{%
Nemanja Hranisavljevic$^1$ \and Oliver Niggemann$^{1,2}$ \and Alexander Maier$^2$\\
$^1$inIT - Institute Industrial IT, Lemgo, Germany\\
e-mail: \{nemanja.hranisavljevic@stud.hs-owl.de, oliver.niggemann@hs-owl.de\}\\
$^2$Fraunhofer Application Center Industrial Automation IOSB-INA, Lemgo, Germany\\
e-mail: \{oliver.niggemann, alexander.maier\}@iosb-ina.fraunhofer.de
}

\date{October 2016\\
Conference: The 27th International Workshop on Principles of Diagnosis: DX-2016At: Denver, Colorado}

\begin{document}

\maketitle

\begin{abstract}
Performing anomaly detection in hybrid systems is a challenging task since it requires analysis of timing behavior and mutual dependencies of both discrete and continuous signals. Typically, it requires modeling system behavior, which is often accomplished manually by human engineers. Using machine learning for creating a behavioral model from observations has advantages, such as lower development costs and fewer requirements for specific knowledge about the system.
  
The paper presents \dad{}, new approach for automatic model learning and anomaly detection in hybrid production systems. It combines deep learning and timed automata for creating behavioral model from observations. The ability of deep belief nets to extract binary features from real-valued inputs is used for transformation of continuous to discrete signals. These signals, together with the original discrete signals are than handled in an identical way. Anomaly detection is performed by the comparison of actual and predicted system behavior. The algorithm has been applied to few data sets including two from real systems and has shown promising results.

\end{abstract}

\section{Introduction}
\label{sec:Introduction}

The analysis and diagnosis of complex production plants has gained new attention due to research agendas such as Cyber-physical Production Systems (CPPSs) \cite{4519604,Rajkumar:2010:CSN:1837274.1837461}, the US initiative Industrial Internet \cite{evans:2012a} or its German pendant ``Industrie 4.0''. In these agendas, a major focus is on the self-diagnosis capabilities for complex and distributed CPPSs. Typical goals of such self-diagnosis approaches are the detection of anomalies, of suboptimal energy consumption, of error causes or of  wear \cite{6059056,isermann:2004a,NL15}. 

Typical challenges for the diagnosis of CPPSs are the increasing complexity of plants, the high frequency of plant modifications---mainly due to product variants---and non-trivial causalities and timing dependencies. A typical CPPS solution for these challenges is a data-driven and machine learning based approach, which complements the traditional solutions based on manual models \cite{Niggemann:2015}. This places data analysis and machine learning algorithms at the heart of CPPS, Industrial Internet and Industrie 4.0. The underlying research question could be stated as follows: Can we extract models of plant behavior, of causalities and of timings, from measured data? Once such models have been learned, anomalies can be identified easily by comparing observations to the learned normal, error causes can be computed by back-tracing the causalities leading to the observations or suboptimal energy consumption can be detected by comparing energy consumption at different operating points.  

So far, a large number of heterogenous machine learning algorithms have been used for this: e.g. state machines are learned for discrete variables such as control signals \cite{niggemann:2012b}, time series have been learned for the timing behavior of continuous signals \cite{WJN2015} or statistical models have been learned for dependent variables \cite{ELG15}. However, in all these cases, discrete and continuous signals heve been analyzed separately and differently, rendering a holistic signal analysis impossible. Even if hybrid automata are learned \cite{niggemann:2012b}, the states will still be defined by discrete signals and continuous signals will be dealt with separately within the states. 

Here, the new algorithm \dad{} is introduced which handles discrete and continuous signals in a uniform way and is able to analyze the system's overall timing. The main idea of the algorithm is shown in an abstract manner in figure \ref{dad1.pdf} (details are given in Section \ref{sec:new_algorithm}): 
\bsfigure[h]{0.4}{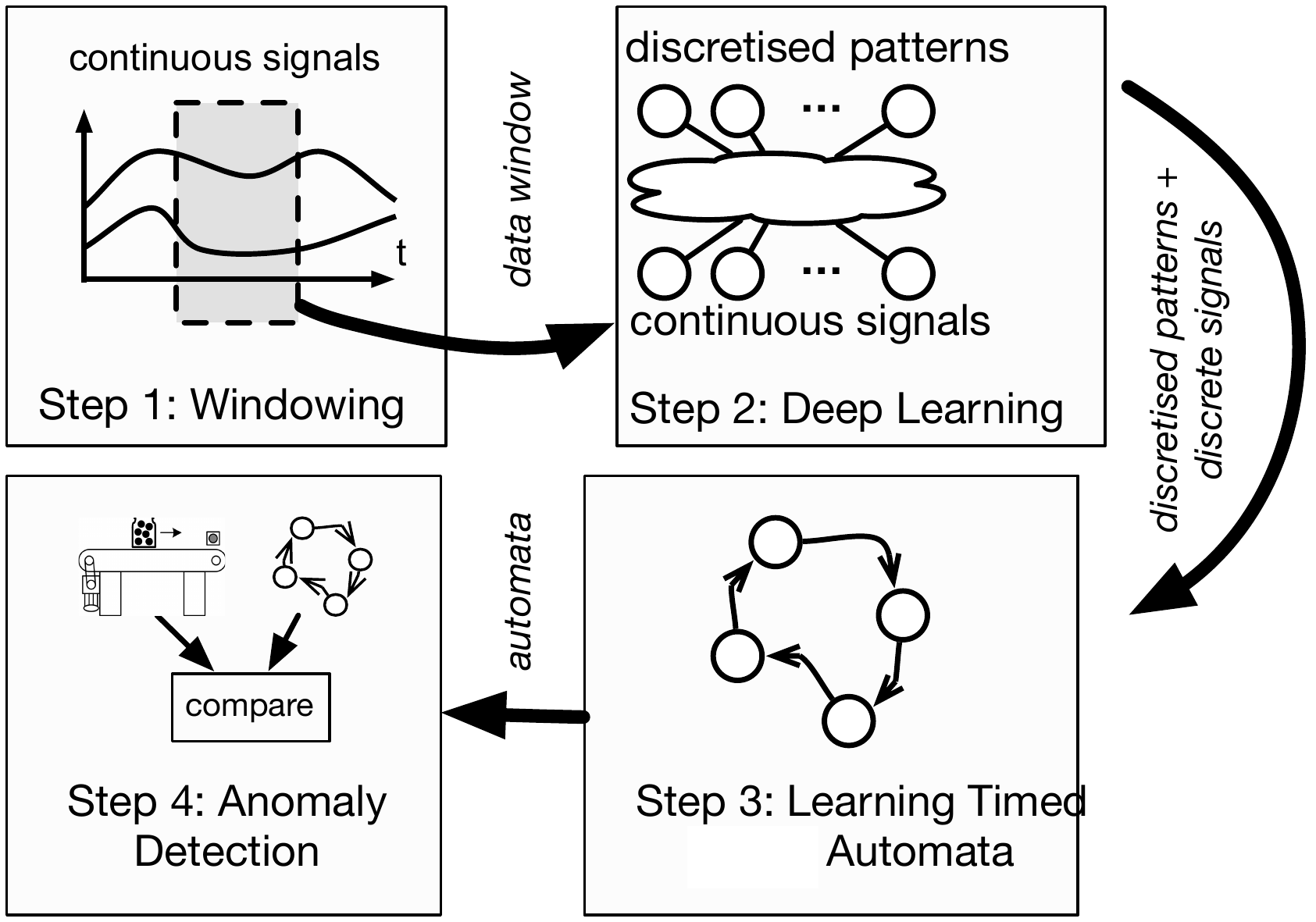}{The main idea of \dad{}}
First, a sliding time window of the set of continuous data is used as the input layer of a deep learning network (step 1). In step 2, this network learns reoccurring patterns in the data. These reoccurring patterns result in a specific binary pattern on the top layer of the deep learning network, i.e. the network discretizes the continuous signals.  Now the observations can be interpreted as a sequence of discrete patterns or events---and can be combined with the discrete signals from the observations. In step 3, from these sequences of patterns or events a timed automata is learned. At this point, we can handle the set of all signals in a uniform discrete manner, this corresponds to a human operator who understands a system behavior as a causal model of reoccurring patterns. Finally, in step 4, anomalies can be detected by comparing the predictions of the learned automata with the actual observations. Typical anomalies are unusual timings, unexpected events or new patterns.

The main contribution of this paper is threefold. First of all, for the first time (to the knowledge of the authors), a deep learning network is used for unsupervised anomaly detection of production systems. Second, based on the deep learning network, a new anomaly detection algorithm \dad{} is introduced which handles continuous and discrete signals in a uniform manner. Third, the approach is evaluated using both real systems and artificial data.

The paper is structured as follows: Section \ref{sec:pre} gives preliminary definitions. In Sections \ref{sec:deep_learning} and \ref{sec:timed_automata} we give an overview and formulate concepts from deep learning and timed autamata theory that are referred to in Section \ref{sec:new_algorithm}, where the outline and formal definition of the new algorithm is presented. Section \ref{sec:experiments} presents some experimental results, while the paper is concluded in Section \ref{sec:conclusion}.

\section{Preliminary definitions} \label{sec:pre}
In this section we give several definitions which we will refer to later in the paper.

\begin{definition}[Signal vector] \label{def:signalVector}
Signal vector is a vector $\textbf{u}=(u_1,  u_2, ..., u_{|\textbf{u}|})^{T}$, where each $u_i|_{i\in\{1,...,{|\textbf{u}|}\}}$ is a value of a discrete ($u_i\in\{0,1\}$) or continuous ($u_i\in\real$) signal and ${|\textbf{u}|}\in\nat$ is a number of signals in an observed system.
\end{definition}

\begin{definition}[Observation example] \label{def:observationExample}
An observation example is a set of pairs $O = \{(t_1,\textbf{u}_1), (t_2,\textbf{u}_2), ... , (t_{|o|},\textbf{u}_{|o|})\}$, where $\textbf{u}_i|_{i\in\{1,..,|o|\}}$ is a signal vector according to Definition \ref{def:signalVector}, $t_i\in\real_{\geq0}|_{i\in\{1,..,|o|\}}$ is the corresponding time stamp and $|o|\in\nat$ is the number of observations in $O$.
\end{definition}

\begin{definition}[Snapshot]  \label{def:snasphot}
A snapshot is a column vector $\textbf{s}$, result of windowing method applied on observation example $O$ (Definition \ref{def:observationExample}). Windowing of observation example $O$ at moment $t_w\in\real_{\geq0}$ using window of length $|\textbf{s}|\in\nat$ and sample time $T_s\in\real_{\geq0}$ gives snapshot $\textbf{s}=(\textbf{u}_1^T||...||\textbf{u}_{|\textbf{s}|}^T)^T$, where $\textbf{u}_i|(t_i,\textbf{u}_i)\in O,t_i=t_w+(i-|\textbf{s}|)\cdot T_s,i\in\{1,...,|\textbf{s}|\}$. $||$ is a symbol for vector concatenation. 
\end{definition}


\section{Deep learning}
\label{sec:deep_learning}
Deep learning refers to a class of machine learning algorithms used to learn deep architectures, where depth of an architecture refers to the number of levels of composition of nonlinear operations in the function learned by the model. Deep learning among other methods, includes training multilayer perceptrons, deep belief nets and deep auto-encoders. 


Typically, deep architectures transform raw input representation of data into gradually more and more abstract representations. We consider that modeling time behavior and mutual dependencies of a large number of signals in a production plant can be done efficiently using deep architectures. Furthermore, we want to make use of often available, large data sets of observations from production plants and find compressed binary representation of plant condition in time.  \dad{} algorithm relies on a restricted Boltzmann machine and deep belief net which will be described in the following sections.

\subsection{Restricted Boltzmann machine}
A restricted Boltzmann machine (RBM) is an undirected probabilistic graphical model with two layers of units (stochastic variables): visible (input) $\textbf{v}$ and hidden $\textbf{h}$ as illustrated on figure \ref{fig:rbm_dbn}. RBM can learn a model of data distribution from training examples presented to its inputs and it is a building block of deep belief net (Section \ref{sec:dbn}). There are different variants of RBM, while we use Bernoulli-Bernoulli (BBRBM) and Gaussian-Bernoulli restricted Boltzmann machine (GBRBM):

\begin{figure}[h]
\centerline{\includegraphics[width=3.3in]{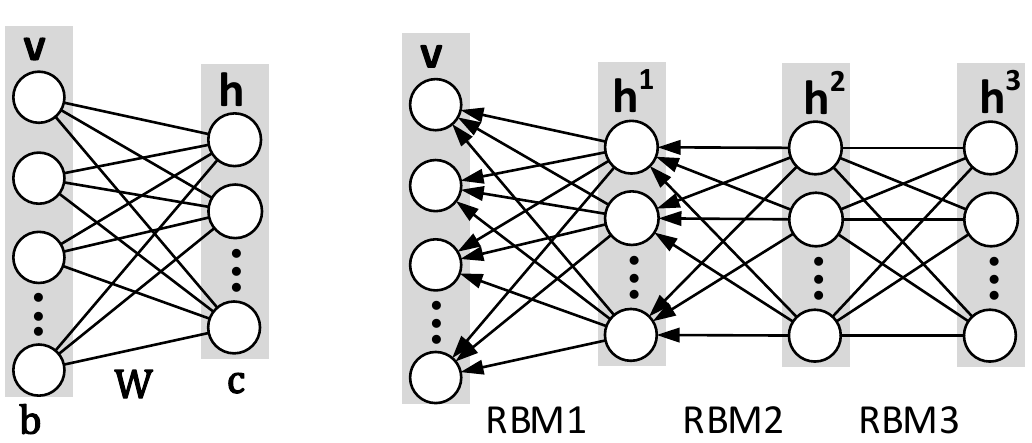}}
\caption{Undirected graphical model of RBM (left) and DBN with three hidden layers (right)---connections between units show their specific conditional dependence structure}
\label{fig:rbm_dbn}
\end{figure}

\begin{definition}[Restricted Boltzmann machine]
\label{def:rbm}
A RBM is an undirected graphical model that models joint probability distribution of visible units $\textbf{v} = (v_1,...,v_m)^T$ and hidden units $\textbf{h} = (h_1,...,h_n)^T$.  It is a tuple $\mathcal{RBM} = (W, b, c, p(\textbf{h}|\textbf{v}), p(\textbf{v}|\textbf{h}))$, where:
\begin{itemize}
\item $W\in\real^{m \times n}$ is weight matrix with each element $W_{i,j}$ describing dependence between $v_i$ and $h_j$
\item $b\in\real^m$ is column vector with each element $b_i$ representing bias of visible unit $v_i$
\item $c\in\real^n$ is column vector with each element $c_j$ representing bias of hidden unit $h_j$
\item $p(\textbf{h}|\textbf{v})$ is probability function of hidden given visible units. $\{h_j|_{j\in\{1,...,n\}}\}$ are conditionally independent given $\textbf{v}$. $p(h_j|\textbf{v})$ is parameterized by $W$, $c$ and is member of exponential family distributions.
\item $p(\textbf{v}|\textbf{h})$ is probability function of visible given hidden units. $\{v_i|_{i\in\{1,...,m\}}\}$ are conditionally independent given $\textbf{h}$. $p(v_i|\textbf{h})$ is parameterized by $W$, $b$ and is member of exponential family distributions.
\end{itemize} 
\end{definition}

\begin{definition}
\label{def:bbrbm}
(Bernoulli-Bernoulli RBM) A BBRBM is a RBM for which:
\begin{itemize}
\item $\textbf{v}$ takes values from $\{0,1\}^m$
\item $\textbf{h}$ takes values from $\{0,1\}^n$
\item Conditional expectation of hidden given visible layer is $\mu^h=E[\textbf{h}|\textbf{v}] = \sigma(c + W^{T}\textbf{v})$ and $p(h_j|\textbf{v})=\mathcal{B}(\mu^h_j)$.\footnote{\label{foot:sigma}$\sigma$ is a sigmoid function defined with $\sigma(x)=(1+e^{-x})^{-1}$}\textsuperscript{,}\footnote{\label{foot:B}$\mathcal{B}(\mu)$ denotes Bernoulli distribution with mean $\mu$}

\item Conditional expectation of visible given hidden layer is $\mu^v=E[\textbf{v}|\textbf{h}] = \sigma(b + W\textbf{h})$ and $p(v_i|\textbf{h})=\mathcal{B}(\mu^v_i)$.
\end{itemize} 
\end{definition}

\begin{definition}
\label{def:gbrbm}
(Gaussian-Bernoulli RBM) A GBRBM is a RBM for which:
\begin{itemize}
\item $\textbf{v}$ takes values from $\real^m$
\item $\textbf{h}$ takes values from $\{0,1\}^n$
\item Conditional probability of hidden given visible layer is defined same as in Definition \ref{def:bbrbm}.
\item Conditional expectation of visible given hidden layer is $\mu^v=E[\textbf{v}|\textbf{h}] = b + W\textbf{h}$ and $p(v_i|\textbf{h})=\mathcal{N}(\mu^v_i, 1)$.\footnote{\label{foot:N}$\mathcal{N}(\mu, 1)$ denotes Gaussian distr. with mean $\mu$ and variance 1}
\end{itemize} 
\end{definition}

In previous definitions we do not give the expressions for joint probability functions of BBRBM and GBRBM, however they can be derived from given conditional probabilities. RBM is an energy based model and its joint distribution is defined through an energy function: function that is mapping each possible configuration of stochastic variables (visible and hidden) to scalar energy. In case of BBRBM the energy function is: $E(\textbf{v},\textbf{h})=b^T\textbf{v}-c^T\textbf{h}-\textbf{v}^TW\textbf{h}$, while for GBRBM it is: $E(\textbf{v},\textbf{h})=\frac{1}{2}(\textbf{v}-b)^T(\textbf{v}-b)-c^T\textbf{h}-\textbf{v}^TW\textbf{h}$. The joint probability function is then $p(\textbf{v},\textbf{h})=e^{-E(\textbf{v},\textbf{h})}/Z$, where $Z$ represents a normalization constant, also called partition function. It is ensuring that total probability for all possible configurations (states) of visible and hidden units is 1 and $Z=\sum_{\textbf{v}}\sum_{\textbf{h}}e^{-E(\textbf{v},\textbf{h})}$ for BBRBM or $Z=\int_{\textbf{v}}\sum_{\textbf{h}}e^{-E(\textbf{v},\textbf{h})}$ for GBRBM. Calculating $Z$ and exact probability (considering that evaluation of $Z$ is required for calculating probability) for specific configuration is intractable in practice. This is because the sums (sum and integral) in expression for $Z$ run through all possible configurations.

Notice that expressions for conditional expectations in Definitions \ref{def:bbrbm} and \ref{def:gbrbm} are defined over $\real^m(\real^n)$. This enables us to expand domain of binary unit to $[0,1]$ (instead of $\{0,1\}$) when we calculate  expectations of other units condition on it---the value from $(0,1)$ is than interpreted as probability of that unit having value 1.
 

\subsubsection{Gibbs sampling}
It is possible to sample from a joint distribution modeled by RBM, which makes it generative model. A procedure that is often used for generating samples is Gibbs sampling (lines 1-4 in Algorithm \ref{alg:cd}). It is an iterative Markov chain Monte Carlo procedure that in each iteration samples each stochastic unit conditioned on the latest values of other units---this makes it efficient for RBM, since sampling all units of one layer (given another) can be done at once. These samples constitute a Markov chain, whose stationary distribution corresponds to the modeled joint distribution. Performing more steps (iterations) leads to more precise samples.

\subsubsection{Training RBM}
Training RBM is a process of learning weights and biases of RBM so it models some data distribution. Simplified, the training algorithm should adjust parameters ($W$, $b$, $c$ from Definition \ref{def:rbm}) of a model so that probability for examples from the training set increases. RBM is trained using stochastic steepest ascent for maximizing the log likelihood of training data under the model, since finding the parameters analytically is not generally possible. 

Contrastive divergence gives us an update rule for training RBM using stochastic steepest ascent. It is approximating derivative of the log likelihood of training set with respect to the parameters ($W$, $b$ or $c$). The result of this approximation is very simple rule (Algorithm \ref{alg:cd}) for updating parameters of RBM where input example is presented to the visible layer of RBM and the parameters are updated in order to increase the probability of that example. Contrastive divergence is denoted CD-$k$, where $k$ is the number of Gibbs steps used.

\begin{algorithm}
	\caption{Contrastive divergence (CD-$k$) update rule, $k \in \nat$, learning rate $\epsilon \in \real^+$}\label{alg:cd}
	\begin{algorithmic}[1]
		\Statex \textbf{In1:} $\mathcal{RBM} = (W, b, c, p(\textbf{h}|\textbf{v}), p(\textbf{v}|\textbf{h}))$ from Definition \ref{def:rbm}
		\Statex \textbf{In2:} Training input $v_0\in\real^m, m\in\nat$
		\Statex \textbf{Out:} Parameters $W$,$b$ and $c$ of $\mathcal{RBM}$ are updated		
		\For {$i = 1...k$}
			\State $h_{i-1} \gets$ Sample from $p(\textbf{h}|\textbf{v}=v_{i-1})$ of $\mathcal{RBM}$
			\State $v_i \gets$ Sample from $p(\textbf{v}|\textbf{h}=h_{i-1})$ of $\mathcal{RBM}$
		\EndFor
		\State $h_0 \gets E[\textbf{h}|\textbf{v}=v_0]$ of $\mathcal{RBM}$ (Definition \ref{def:rbm})
		\State $h_k \gets E[\textbf{h}|\textbf{v}=v_k]$ of $\mathcal{RBM}$ (Definition \ref{def:rbm})
		\State $W \gets W + \epsilon(v_0h_0^T - v_kh_k^T)$
        \State $b \gets b + \epsilon(v_0 - v_k)$ 
        \State $c \gets c + \epsilon(h_0 - h_k)$
	\end{algorithmic}
\end{algorithm}

Previously described training procedure allows introduction of several modifications/additions to improve the learning process and the final model. Steepest ascent alows us to configure: batch processing of training examples, the number of epochs/iterations used. Contrastive divergence rule can be modified by: introduction of momentum, sparsity of hidden units, weight decay. It is also important to select proper type of RBM, number of hidden units, initial weights and biases. \cite{Hinton:2012}

\subsection{Deep belief net and greedy layer-wise training}
\label{sec:dbn}
This section describes deep belief net, a deep architecture used in this paper. The model and a learning algorithm are introduced in \cite{Hinton:2006} and have been applied to many different tasks, some of which are classification, regression, natural language processing, information retrieval \cite{Bengio:2009}.

Deep belief net (DBN) can be used for finding low-dimensional representation of an input \cite{Hinton:2011} by gradually extracting more and more abstract representation. These higher representations are expressed by activation of some subset of units (feature detectors) that are not mutually exclusive---they form a "distributed representation". Using DBN it is possible to extract binary features from raw continuous data (e.g. signal observations)---in our case we are interested in finding binary representations of reoccurring patterns in continuous signals from an industrial system. We want to find short binary representation (code) whose each unit (bit) captures some, possibly important, structure of the input data. Since each unit can have value of 0 or 1, it is separating input space on two regions. The whole code corresponds to some small region of the input space---all inputs from this region are "similar" from a perspective of DBN, as they activate the same units (feature detectors).  

\textit{Example}: Suppose we have a system with two drives used for movement of some objects left/right and up/down. Energy consumption of each of the drives is observed. Binary coding of these real-valued observations might give us 3 binary values whose interpretation could be: object is moved up/down, object is moved left/right and object is light/heavy (supposing that there are two different weight classes of the moved objects). It is usually not possible to interpret these individual binary values, except for very simple models.

DBN with $l\in\nat$ hidden layers is a deep model whose top two layers ($\textbf{h}_l$, $\textbf{h}_{l-1}$) form a restricted Boltzmann machine and lower layers ($\textbf{v}$,$\textbf{h}_1$,...,$\textbf{h}_{l-2}$) form a directed graphical model---stochastic neural network in which units in each layer are independent, given the values of the units in the layer above. DBN is illustrated in figure \ref{fig:rbm_dbn} and formally defined in Definition \ref{def:dbn}.

\begin{definition}
\label{def:dbn}
(Deep belief net) A DBN with an architecture $m$-$n_1$-...-$n_l$ models joint probability distribution $p(\textbf{v}, \textbf{h}^1,...,\textbf{h}^l)$ where $\textbf{v}=(v_1,...,v_m)$ is a vector (layer) of visible (input) units and $\textbf{h}^j=(h^j_1,...,h^j_{n_j})|_{j\in\{1,...,l\}}$ are $l$ vectors (layers) of hidden units, $m,n_1,...n_l,l \in \nat$. It is a sequence $\mathcal{DBN} = (\mathcal{RBM}_j)_{j \in \{1,...,l\}}$, where:
\begin{itemize}
\item $\mathcal{RBM}_j$ is RBM from Definition \ref{def:rbm}

\item $p(\textbf{v}, \textbf{h}^1,...,\textbf{h}^l)=p(\textbf{h}^{l-1}, \textbf{h}^l)(\prod_{k=1}^{l-2}p(\textbf{h}^k|\textbf{h}^{k+1}))p(\textbf{v}|\textbf{h}^1)$ is a joint probability function

\item $p(\textbf{h}^{l-1}, \textbf{h}^l)$ in equation for $p(\textbf{v}, \textbf{h}^1,...,\textbf{h}^l)$ is joint probability function of $\mathcal{RBM}_l$.

\item $p(\textbf{h}^j|\textbf{h}^{j+1}))|_{j=1,...,l-2}$ in equation for $p(\textbf{v}, \textbf{h}^1,...,\textbf{h}^l)$ is conditional probability function of visible given hidden units of $\mathcal{RBM}_j$ from $\mathcal{DBN}$.

\item $p(\textbf{v}|\textbf{h}^1)$ in equation for $p(\textbf{v}, \textbf{h}^1,...,\textbf{h}^l)$ is conditional probability function of visible given hidden units of $\mathcal{RBM}_1$ from $\mathcal{DBN}$.

\item $\textbf{v}$, $\textbf{h}^j|_{j\in\{1,...,l\}}$ take values from the sets def. by $\mathcal{RBM}_j$
\end{itemize} 
\end{definition}


\subsubsection{Training DBN}
In \cite{Hinton:2006}, greedy layer-wise training algorithm for training $l$-layer DBN from observations is presented, which sequentially trains $l$ RBMs using contrastive divergence. The procedure is described in Algorithm \ref{alg:train_dbn}. Each of the RBMs corresponds to one level of DBN. The input training set is used for training the first (lowest) RBM, while the training set for each next RBM is obtained by transforming the training set of the RBM below it, as described in Algorithm \ref{alg:train_dbn}.

\begin{algorithm}
	\caption{Training deep belief net, $l\in\nat$ is number of hidden layers, meta parameters $dbn\_mp$ for training each RBM are provided}\label{alg:train_dbn}
	\begin{algorithmic}[1]
		\Statex \textbf{In:} Training examples  $\mathcal{X}=(x_i)_{i\in \{1,...,|\mathcal{X}|\}}$, where $|\mathcal{X}|\in\nat$ is number of examples and $x_i\in\real^m$, $m\in\nat$
		\Statex \textbf{Out:} Learned model $\mathcal{DBN} = (\mathcal{RBM}_j)_{j \in \{1,...,l\}}$ (Def. \ref{def:dbn})
		\State $\mathcal{X}_1 \gets \mathcal{X}$
		\For {$j=1,...,l$}
			\State $\mathcal{RBM}_j \gets$ Train $\mathcal{RBM}_j$ using steepest ascent and CD-k (Algorithm \ref{alg:cd}) with training examples from $\mathcal{X}_j$
			\State $\mathcal{X}_{j+1} \gets$ Transform each example $x_i$ from $\mathcal{X}_j$ using expectation of hidden given visible layer $E[\textbf{h}|\textbf{v}=x_i]$ of previously learned $\mathcal{RBM}_j$ (see Definitions \ref{def:rbm}, \ref{def:bbrbm}, \ref{def:gbrbm})
		\EndFor
	\end{algorithmic}
\end{algorithm}

Theoretical justification of training algorithm and introduction of multiple levels in DBN is not subject of this paper. Refer to \cite{Hinton:2006,Bengio:2009} for more information.

\begin{algorithm}
	\label{dbn_transform}
	\caption{DBN - Transformation of input vector to top level binary representation}\label{alg:dbn_transform}
	\begin{algorithmic}[1]
		\Statex \textbf{In1:} Input vector $v^1\in\real^m,m\in\nat$
		\Statex \textbf{In2:} $\mathcal{DBN} = (\mathcal{RBM}_j)_{j \in \{1,...,l\}}$ (Definition \ref{def:dbn})
		\Statex \textbf{Out:} Binary code $b\in\{0,1\}^{n_l}, n_l\in\nat$ is number of units at top layer of DBN
		\For {$j=1,...,l$}
			\State $h^j \gets$ Expectation of hidden given visible layer of $\mathcal{RBM}_j$ from $\mathcal{DBN}$: $E[\textbf{h}|\textbf{v}=v^j]$ (see Definitions \ref{def:rbm}, \ref{def:bbrbm}, \ref{def:gbrbm})
			\State $v^{j+1} \gets h^j$
		\EndFor
		\State $b \gets$ Round each value from $h_l$ to nearest of \{0,1\}
	\end{algorithmic}
\end{algorithm}
\subsubsection{Transforming input to top level binary representation}
Modeling data distribution using DBN results in extracting higher level features from row input data \cite{Hinton:2011}. Therefore, training DBN with Gaussian-Bernoulli RBM at the input layer and Bernoulli-Bernoulli RBMs at other layers enables us to transform real-valued input vector to top level binary representation.
This procedure is described in Algorithm\ref{alg:dbn_transform}.

Expected value of each unit at top layer of DBN ($\textbf{h}_l$ in Algorithm \ref{alg:dbn_transform}) gives us the unit probability of having value 1. Each of these probabilities can be interpreted as intensity of input vector having some structure detected by corresponding unit. We round them to 0 or 1 to get binary representation of given input vector. Similarly, it is possible to propagate the binary code from the top layer back to the input layer. Hence, we are reconstructing the input from it's code. 



\section{Timed Automaton}
\label{sec:timed_automata}

The identification of behavior models for discrete signals requires different techniques than for continuous signals.
A finite state automaton (FSA) is an established formalism which is able to capture the discrete behavior.
Here, additionally the timing information is considered and therefore a timed automaton (TA) is identified.



\begin{definition}[Timed Automaton] \label{def:TA}
A timed automaton is a 4-tuple $A = (S, \Sigma, T, \delta)$, where 
\begin{itemize}
\item $S$ is a finite set of states. Each state $s \in S$ is a tuple $s = (id, \textbf{u})$, where $id$ is a current numbering and $\textbf{u}$ is a discrete signal vector according to Definition \ref{def:signalVector}.

\item $\Sigma$ is the alphabet, the set of events.
	\item $T$ is a set of transitions. A transition 
is represented with $(s, a, \delta , s')$, where $s, s' \in S$ are the source and destination states, $a\in \Sigma$ is the symbol and $\delta$ is a clock constraint. 
The automaton changes from state $s$ to state $s'$ triggered by a symbol $a \in \Sigma$ if the current clock value satisfies $\delta$.
The clock $c$ is set to $0$ after executing a transition, so that the clock starts counting time from executing this transition. 
\item A transition timing constraint $\delta: T \rightarrow I$, where $I$ is a set of intervals. 
$\delta$ always refers to the time spent since the last event occurred. It is expressed as a time range or as a probability density function (PDF), i.e. as probability over time.
\end{itemize}
\end{definition}


The OTALA timed automaton learning algorithm uses the information from the signal vector to extract state information.
The event between two states is created as the difference between the corresponding state vectors, additionally including the timing conditions (e.g. time range between minimum and maximum observation time or probability density function (PDF) over time). An outline of the algorithm OTALA is given in Algorithm \ref{alg:otala}, for a complete version refer to \cite{Maier:2014a}.

\begin{algorithm}[h]
	\caption{Timed automaton learning algorithm}\label{alg:otala}
	\begin{algorithmic}[1]
		\Statex \textbf{In:} Observation example $D$ according to Definition \ref{def:observationExample}  
		\Statex \textbf{Out:} Timed automaton $\mathcal{A}$ according to Definition \ref{def:TA}
		\While {(identificationConverged()==$false$)}
			\State $t,\textbf{u}\gets$ getNextEvent($D$)
			\If{stateExists($s,\textbf{u}$)}
					\If{transitionExists($s_{current},s$)}
						\State updateStateInformation($s$)
						\State updateTransitionInformation($s_{current},s,t$)
					\Else 
						\State createNewTransition($s_{current},s$)
						\State updateStateInformation($s$)
					\EndIf
			\Else 
					\State createNewState($s,\textbf{u}$)
			\EndIf
			\State $s_{current} \gets s$
		\EndWhile
		\State \Return $\mathcal{A}$
	\end{algorithmic}
\end{algorithm}

Simplified, the Algorithm \ref{alg:otala} works as follows:
For each incoming learning sample (vector of discrete signals including the time stamp), it is checked whether a corresponding state and transition exists (lines 3-5).
If they already exist, they are updated (timing information, lines 5-12), otherwise they are created (lines 9-15).
The identification runs until convergence (checked in lines 17-18).

Anomaly detection is done by comparing new observations with the behavior of the identified automaton.
Generally, the behavior of a system can be described as a path through the automaton.
As long as this path can be followed, the observed behavior corresponds to a normal behavior.
Whereas, as soon as some observation cannot be mapped to the automaton, an anomaly is detected.

The anomaly detection procedure based on timed automata is first published in \cite{VKBNM2011b} and is outlined in Algorithm \ref{alg:ad}. The algorithm is based on the validation of incoming events.
Therefore, in line 3 an event is extracted from the observation sample.
The corresponding event is the difference between two subsequent signal vectors.
In line 4, it is checked whether the event is available in the automaton, respectively, in line 5, the timing is checked.
If an anomaly is found, it is returned in line 8 and 11 respectively, otherwise the anomaly detection is continued with the next state in line 6.

\begin{algorithm}[htb!]
	\caption{Anomaly Detection Algorithm}\label{alg:ad}
	\begin{algorithmic}[1]
		\Statex \textbf{In1:} Timed automaton $\mathcal{A}$ (Definition \ref{def:TA})
		\Statex \textbf{In2:} Observation example $D$ according to Definition \ref{def:observationExample}
		\Statex \textbf{Out:} Detected anomaly(if there exists one): $anomaly$ 
		\State $s \gets s_0$, state is assigned to be the initial state $s_0$
		\ForAll{tuples in $D$}
			\State $e\gets$extractEventFromObservation($D$)
			\If{exists $e \in T$} 
				\If{$t$ satisfies $\delta(T)$} 
					\State $s\gets s'$ 
				\Else
				\State \Return{$anomaly$: wrong timing}
				\EndIf
			\Else
				\State \Return{$anomaly$: unknown event}
			\EndIf
		\EndFor
	\end{algorithmic}
{\small
}

\end{algorithm}

\section{\dad{} algorithm} \label{sec:new_algorithm}
In this section we will outline the basic idea behind the proposed algorithm using a simple example and later we will formally define it.

\subsection{Algorithm outline}
\label{sec:Algorithm_outline}
\dad{} algorithm is outlined using artificial data set. In section \ref{sec:Introduction}, four steps of an algorithm are declared. Each of the steps will be described in more details using example data.

\begin{figure}[h]
\centerline{\includegraphics[width=3.3in]{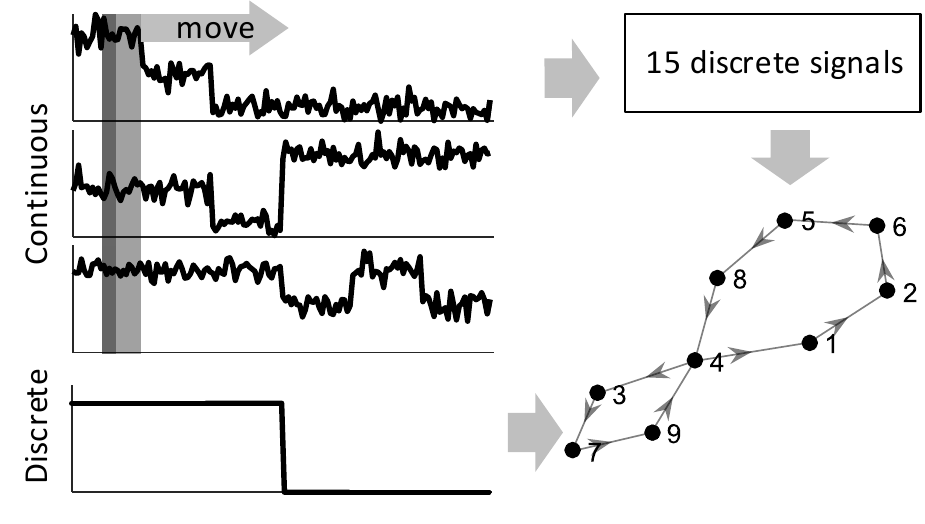}}
\caption{\dad{} applied on artificial data} \label{fig:outline_config2}
\end{figure}

\textit{Example data set:} We have generated a simple artificial time-series data set with four signals: three continuous and one discrete. Data set contains observations during 200 system cycles, obtained by modifying "base cycle". The sequence of this cycle is divided in six sub-sequences during each of which signals have constant values. Performed modifications of base signals include adding Gaussian noise to continuous signals and making small differences in duration of sub-sequences. One generated cycle is illustrated on left half of Figure \ref{fig:outline_config2}. 

\textit{Step 1: Windowing - }
One system cycle is time-series data over 30s at 150 time moments. The snapshots (Definition \ref{def:snasphot}) are obtained using 3s window---a snapshot contains 15$\cdot$3 real values (15 observations for each of the three continuous signals). Window overlapping of 30\% was used which resulted in 2800 snapshots in deep learning training set.


\textit{Step 2: Deep Learning - }
In this step deep belief net is trained. Identifying the right meta parameters of DBN is usually done manually, based on experience, and it is not the subject of this paper. However, in this example we used four layer DBN with 15 units in top layer. Similar snapshots (according to DBN) are coded into the same binary value. Binary coding of all snapshots in one cycle and attaching information about the time stamp to each code is "discretizing" the signals, so the result is that we have replaced 3 continuous with 15 discrete signal. Notice that snapshots contain information about signals during 3s and each contains 45 real values.


\textit{Step 3: Learning Timed Automaton - }
15 discrete signals that are result of transformation from continuous signals and one original discrete signal are merged together and used to learn timed automaton. The resulting automaton had nine states and its state diagram is shown in Figure \ref{fig:outline_config2}. 

\textit{Step 4 Anomaly detection - }
Anomaly detection in a sequence of new observations requires continuous observations to be discretized using DBN. The system behavior is then described by change of these discrete signals in time. When these signals are committed to the learned automaton---the path through the automaton is defining system behavior. As long as a path through the automaton can be followed, the observed behavior corresponds to normal behavior. Conversely, as soon as some combination of discrete signals cannot be mapped to the automaton, an anomaly is detected.

Anomalies are detected in one of the situations: 
\begin{itemize}
\item Unknown event in the automaton caused by new pattern of continuous signals.
\item Unknown event in the automaton caused by novel combination of discrete signals (original and transformed)
\item Wrong timing of an event
\item Unknown event caused by novel transition between two learned states 
\end{itemize}

\subsection{Formal definition of a new algorithm}

Formally we will define \dad{} using two procedures: \textit{Learning behavioral model} (Algorithm \ref{alg:learn} and \textit{Anomaly detection} (Algorithm \ref{alg:det_anomaly}). The first one includes: \textit{Windowing}, \textit{Deep learning} and \textit{Learning automaton} steps while the second one includes \textit{Anomaly detection} step. Outline of these steps is given in previous section.

\begin{algorithm}[h]
	\caption{Learning behavioral model, parameters are: window size $T_w\in\real_{\geq0}$, windows overlap $T_o\in\real_{\geq0}$, meta parameters for DBN training $dbn\_mp$ }\label{alg:learn}
	\begin{algorithmic}[1]
		\Statex  \textbf{In:} Observations $\mathcal{O}=\{O_1,...,O_{|\mathcal{O}|}\}, |\mathcal{O}|\in\nat$ where $O_i|_{i\in\{1,...,|\mathcal{O}|\}}$ is an observation example (Definition \ref{def:observationExample})
		\Statex  \textbf{Out1:} $\mathcal{DBN} = (\mathcal{RBM}_j)_{j \in \{1,...,l\}},l\in\nat$ (Definition \ref{def:dbn})
		\Statex \textbf{Out2:} Timed automaton $\mathcal{A}$ (Definition \ref{def:TA})
		\newline
				
		\ForAll{$O_i\in\mathcal{O}$}
			\State $S_i \gets$ Apply windowing to continuous signals from $O_i$ using rectangular window of length $T_w$ and overlap $T_o$, $S_i=\{(t_k,s_k)|_{k\in\{1,...,|S_i|}\}$ where $s_k$ is a snapshot obtained at moment $t_k$, and $|S_i|$ is the number of obtained snapshots 
		\EndFor
		
		\State $\mathcal{DBN} \gets$ Learn deep belief (Algorithm \ref{alg:train_dbn}) using sequence $S=(s)|_{(\cdot,s)\in\ S_i,i={1,...,|S_i|}}$ as input and $dbn\_mp$		
		
		\ForAll{$S_i\in\mathcal{S}$}
			\State $D_i \gets$ Transform continuous signals from $S_i$ (Algorithm \ref{alg:dbn_transform}) using $\mathcal{DBN}$ and $S_i$ as inputs and append result with original discrete signals from $O_i$
		\EndFor	
		
		\State $\mathcal{A} \gets$ Learn automaton (Algorithm \ref{alg:otala}) using each of observation examples $D_1,...,D_{|D|}, |D|\in\nat$ as inputs
	\end{algorithmic}
\end{algorithm}

In Algorithm \ref{alg:learn} first windowing is performed on each production cycle---observation example (Definition \ref{def:observationExample}). This results in data set of snapshots for training deep belief net. DBN is trained using snapshots from all cycles. Learned DBN model enables us transformation of snapshots to binary codes which is used to transform continuous signals from each cycle to discrete signals: value of some $k$-th discrete signal at moment $t_1$ is value of the $k$-th bit of a binary code of the snapshot covering $t_1$. These discretized signals, together with original discrete signals are used for learning timed automaton.

\begin{algorithm}[h]
	\caption{Anomaly detection, parameters are: window size $T_w\in\real_{\geq0}$, windows overlap $T_o\in\real_{\geq 0}$}\label{alg:det_anomaly}
	\begin{algorithmic}[1]
		\Statex \textbf{In1:} Observation example $O$ (Definition \ref{def:observationExample})
		\Statex \textbf{In2:} Deep belief net $\mathcal{DBN}$ (Definition \ref{def:dbn})
		\Statex \textbf{In3:} Timed automaton $\mathcal{A}$ (Definition \ref{def:TA})
		\Statex \textbf{Out:} Anomaly (if there exist one) $a$
		\newline
		\State $S \gets$ Apply windowing to continuous signals from $O$ using rectangular window of length $T_w$ and overlap $T_o$, $S=\{(t_k,s_k)|_{k\in\{1,...,|S|}\}$ where $s_k$ is a snapshot obtained at moment $t_k$, and $|S|$ is the number of obtained snapshots 
		\State $D \gets$ Transform continuous signals from $S$ (Algorithm \ref{alg:dbn_transform}) using $\mathcal{DBN}$ and $S$ as inputs and append original discrete signals from $O$
		\State $a \gets$ Detect anomaly (Algorithm \ref{alg:ad}), inputs are $\mathcal{A}$ and $D$
	\end{algorithmic}
\end{algorithm}

Algorithm \ref{alg:det_anomaly} describes a simple procedure. For a new observation example, first windowing is applied in the same way as in algorithm \ref{alg:learn} (the same window size and overlap). The sequence of snapshots is than transformed to discrete signals that are together with original discrete signals processed using algorithm \ref{alg:ad}.

\subsubsection{Discussion about \dad{}}
Success of anomaly detection phase depends highly of deep belief net model: \begin{itemize}
\item If DBN does not generalize well, unseen snapshots from the normal behavior will be coded as wrong patterns and false anomaly will be detected. 
\item How is model separating input space on different regions (codes) is not predictable. Binary representation (code) could miss some signal structure that is important for distinguishing between two different patterns
\end{itemize}
We do not give any instructions for measuring quality of learned DBN model, except for using final accuracy score after applying \dad{} on validation data. DBN and timed automata are splitting the "job" here, therefore using different windowing and DBN configuration would result in different automaton. Number of top layer units of DBN, which is related to the number of states of automaton, probably plays most important role in configuring the algorithm. Effects of different configurations of the algorithm are not subject of this paper.

DBN is learning to represent well snapshots from the training set, however it is not predictable how are unseen anomalous snapshots going to be coded. They might be coded to a binary code that never occurred during normal behavior of the system which would automatically imply an anomaly. Another possible effect is that anomalous snapshot will be recognized as one of the patterns that are modeled in automaton, and then whether anomaly will be detected or not depends of learned automaton (transitions and timings). Detecting anomalous patterns is a side effect of the algorithm as it is not intended do detect this type of anomalies.
The targeted anomalies are those with wrong timings of events in the automaton and transition to wrong state. If we assume that any new sequence off snapshots does not contain novel snapshots (patterns), then each snapshot will be coded as proper binary code and automaton will be responsible to detect anomaly---which is done in more predictable way.

\section{Experiments}
\label{sec:experiments}
\subsection {High rack storage system}
\begin{figure}[h]
\centerline{\includegraphics[width=3in]{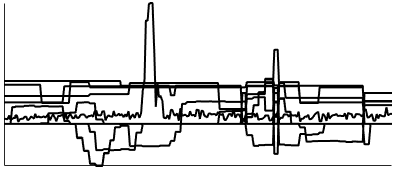}}
\caption{13 continuous signals recorded from High Rack Storage System during one cycle. Signals are normalized to unit variance and ploted on the same axis.} 
\label{fig:storage_system_signals}
\end{figure}

High rack storage system is a demonstrator system for transportation and storage of objects/items. 13 continuous signals are observed in 13065 points during 48 operating cycles. The signals during one cycle are plotted on Figure \ref{fig:storage_system_signals} and they include measured and desired speed of several conveyors and their total energy consumption.

\begin{figure}[h]
\vspace{-10pt}
\centerline{\includegraphics[width=3in]{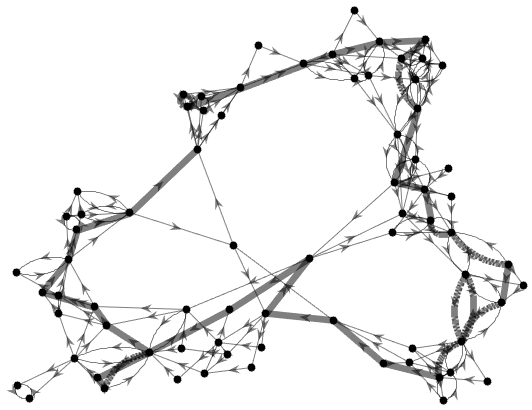}}
\caption{State diagram of the automaton learned from storage system data. Transitions during one of 48 cycles are presented with bold edges.} 
\label{fig:storage_system_graph}
\end{figure}

Deep learning data set was created without windowing, so each training example contained 12 real values (one of the signals was removed because it was constant). Each of the signals is normalized to unit variance. Deep belief net with 3 hidden layers was trained: 12-100-70-40 (the numbers represent number of units at each layer). All the examples from the training set were recognized as one of the 73 different patterns. The model enabled transformation of 12 continuous into 40 discrete signals, that were used for learning (non timed) automaton. State diagram of this automaton is illustrated on Figure \ref{fig:storage_system_graph}.

\begin{figure}[h]
\centerline{\includegraphics[width=3in]{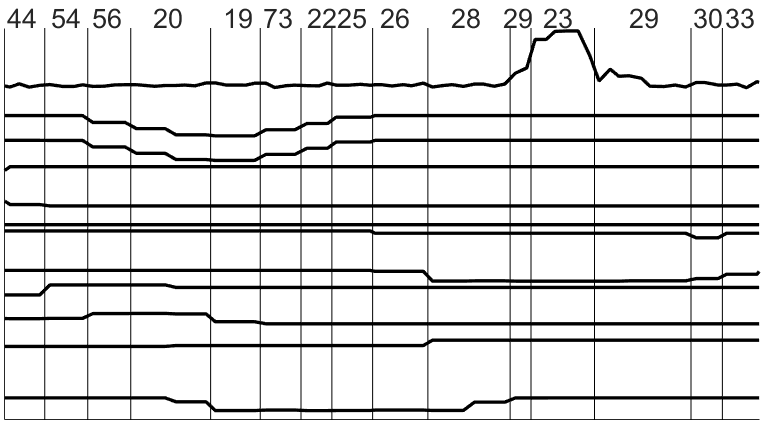}}
\caption{Change of signals and automaton states in time.} 
\label{fig:storage_system_sequence}
\end{figure}

Figure \ref{fig:storage_system_sequence} shows state diagram of the automaton (top of the figure) and signals in time during a part of one production cycle. After visual analysis of plots like this one, during several cycles, we believe that the model really identifies different patterns in input signals. E.g. in figure \ref{fig:storage_system_sequence} there is a peak in top signal---during this peak automaton is in state 23. If the peak is too steep we could get transition from state 28 to 23 which would be detected as an anomaly - unexpected event.

In order to test the algorithm we have artificially modified the signals by adding Gaussian noise with variance 1 at to each of them at some time moment. Table \ref{tab:storage} shows how are different modifications recognized by \dad: as normal behavior, unexpected initial state, new pattern or unknown event.

\begin{table}[h]
\small
\setlength{\tabcolsep}{3pt}
\begin{center}
{\caption{Anomaly detection in artificially modified data from Storage system}\label{tab:storage}}
\vspace{-7pt}
\begin{tabular}{ccccc}
\hline
\rule{0pt}{0pt}
&\multicolumn{4}{c}{Results of anomaly detection}\\
\cline{2-5}
\rule{0pt}{8pt}
Modif.&Unexp. init. st.&New pattern&Unknown event&Normal\\
\hline
\quad$\Diamond$&44\%&40\%&1\%&15\%\\
\quad$\Box$&0\%&63\%&24\%&13\%\\
\hline
\multicolumn{5}{l}{$\Diamond$ Added noise to observations at first time moment in a cycle}\\
\multicolumn{5}{l}{$\Box$ Added noise to observations at random time moment in a cycle}
\end{tabular}
\vspace{-15pt}
\end{center}
\end{table}

\subsection {ATM card reader}
In our second experiment we have used observations of an energy signal of an ATM card reader device/system that is consisted of several energy consuming devices some of which are: electrical motors and electromagnets. Data set contains measurements in 2311629 points during 250 operating cycles.

Deep learning data set was created using window size of 100 samples and 30\% overlap, so each training example contained 100 real values. Deep belief net with 2 hidden layers was trained: 100-60-20. All examples from the training set were recognized as one of the 57 different patterns. The model enabled transformation of 1 continuous (latest 100 observations) into 20 discrete signals, that were used for learning (non timed) automaton. 

\begin{figure}[h]
\centerline{\includegraphics[width=2.7in]{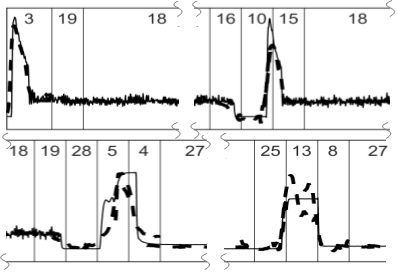}}
\caption{Change of states through time. Part of one cycle of Card Reader energy consumption is shown. Dashed line is reconstruction of snapshot (window overlap of 30\% is used)} 
\label{fig:card_reader_sequence}
\end{figure}

\begin{figure}[h]
\centerline{\includegraphics[width=3in]{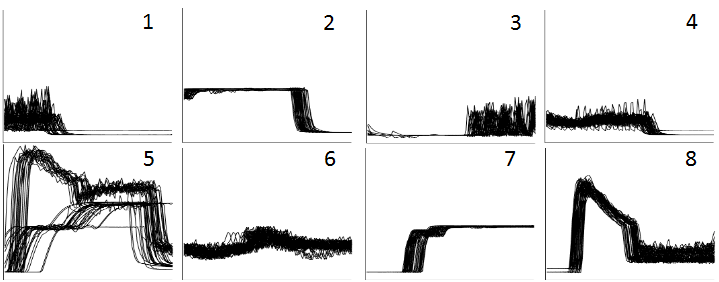}}
\caption{8/57 learned patterns from card reader data---snapshots that are recognized as same discrete pattern are plotted on the same axis.
\label{fig:card_reader_patterns}} 
\end{figure}

Figure \ref{fig:card_reader_sequence} shows energy consumption signal and change of the state of the automaton (printed above the signal) in time during a part of one operating cycle. Parts of the cycle in which there is no change of the state are cut from the plot. In addition, the figure also shows the reconstructions of the input snapshots (obtained by propagating the binary code back to the input layer). Figure \ref{fig:card_reader_patterns} shows patterns learned by deep belief net. After visual analysis of plots like those on Figures \ref{fig:card_reader_sequence} and \ref{fig:card_reader_patterns}, during several cycles, we believe that the model really identifies different patterns in input signal. E.g. in Figure \ref{fig:card_reader_sequence} state 28 is followed by specific patterns recognized as state 5 and than 4. If these patterns are missing (the transition is from 28 to 27) it would be detected as an anomaly - unexpected event. If timed automaton is learned, staying too long in state 28 could be easily detected. 

However, plot of pattern 5 in Figure \ref{fig:card_reader_patterns} shows that different input patterns (classified by human) are sometimes recognized as the same pattern by DBN (pattern 5). This is probably because of the small number of such snapshots in DBN training set and it would result in algorithm failing to detect some anomalies.

In order to test the algorithm we have artificially modified the energy consumption signal by modifying 100 consecutive measurements. Table \ref{tab:storage} shows how are different modifications recognized by \dad: as normal behavior, new pattern or unknown event.

\begin{table}
\small
\setlength{\tabcolsep}{3pt}
\begin{center}
{\caption{Anomaly detection in artificially modified data from Card reader system}\label{tab:card}}
\begin{tabular}{cccc}
\hline
\rule{0pt}{0pt}
&\multicolumn{3}{c}{Results of anomaly detection}\\
\cline{2-4}
\rule{0pt}{0pt}
Modification&New pattern&Unknown event&Normal\\
\hline
\quad$\Diamond$&50\%&45\%&5\%\\
\quad$\Box$&36\%&53\%&11\%\\
\quad$\bigtriangleup$&15\%&35\%&50\%\\
\hline
\\[-6pt]
\multicolumn{4}{l}{$\Diamond$ Signal dropped to zero}\\
\multicolumn{4}{l}{$\Box$ Signal raised by 50\%}\\
\multicolumn{4}{l}{$\bigtriangleup$ Added ramp raising from 0 to the variance of the signal}
\end{tabular}
\vspace{-15pt}
\end{center}
\end{table}

\section{Conclusion}
\label{sec:conclusion}
We have presented a new algorithm that tries to solve the problem of anomaly detection in hybrid production systems. The new algorithm \dad{} relies on deep belief nets and timed automata and minimizes requirements for expert knowledge. The sub procedures for creating behavioral model from observations and comparison of learned model with new observations are given formally. The algorithm was later tested using several data sets. The main purpose of the experiments was to visualize modeled behavior and show different types/classes of detected artificially created anomalies. Many tasks are left for future work some of which are: analysis of an effect of different configuration of deep belief net on accuracy, testing the algorithm using real anomalies from the systems.


\newcommand{\bslitpath}{./literature}

\bibliographystyle{unsrt}
\bibliography{%
\bslitpath/automation-modeling-lit,%
\bslitpath/automation-simulation-lit,%
\bslitpath/cbr-lit,%
\bslitpath/clustering-lit,%
\bslitpath/diagnosis-lit,%
\bslitpath/graph-drawing-lit,%
\bslitpath/graph-lit,%
\bslitpath/energy-lit,%
\bslitpath/learning-dm-lit,%
\bslitpath/logic-lit,%
\bslitpath/math-lit,%
\bslitpath/misc-lit,%
\bslitpath/network-lit,%
\bslitpath/se-lit,%
\bslitpath/maschbau-apps-lit,%
\bslitpath/visualization-lit}
\end{document}